\documentclass[10pt]{article}

\usepackage[T1]{fontenc}
\usepackage[utf8]{inputenc}
\usepackage{amsmath}
\usepackage{newtxtext,newtxmath}
\usepackage[letterpaper,textwidth=5.5in,textheight=9in,centering,footskip=30pt]{geometry}
\usepackage{microtype}
\usepackage{graphicx}
\usepackage{booktabs}
\usepackage{tabularx}
\usepackage{multirow}
\usepackage{array}
\usepackage{xcolor}
\usepackage{enumitem}
\usepackage[numbers,sort&compress]{natbib}
\usepackage{tikz}
\usetikzlibrary{arrows.meta,positioning,fit,backgrounds,calc}
\usepackage{pgfplots}
\pgfplotsset{compat=1.18}
\usepackage{fancyvrb}
\usepackage{caption}
\usepackage[section]{placeins}
\usepackage{flafter}
\usepackage{float}
\usepackage{algorithm}
\usepackage[noend]{algpseudocode}
\usepackage{titlesec}
\usepackage{fancyhdr}
\usepackage[colorlinks=true,linkcolor=blue!50!black,citecolor=blue!50!black,urlcolor=blue!50!black]{hyperref}
\titleformat{\section}{\large\bfseries}{\thesection}{1em}{}
\titlespacing*{\section}{0pt}{12pt plus 3pt minus 2pt}{5pt plus 1pt}
\titleformat{\subsection}{\normalsize\bfseries}{\thesubsection}{1em}{}
\titlespacing*{\subsection}{0pt}{9pt plus 2pt minus 2pt}{3pt plus 1pt}
\titleformat{\paragraph}[runin]{\normalsize\bfseries}{}{0pt}{}
\titlespacing*{\paragraph}{0pt}{5pt plus 1pt}{0.6em}

\fancypagestyle{firstpage}{\fancyhf{}\fancyfoot[L]{\small Preprint.}\fancyfoot[C]{\thepage}}

\renewenvironment{abstract}
  {\begin{center}{\large\bfseries Abstract}\end{center}\vspace{-1.2ex}%
   \begin{list}{}{\setlength{\leftmargin}{0.5in}\setlength{\rightmargin}{0.5in}\setlength{\parsep}{0pt}}\item\relax}
  {\end{list}}

\definecolor{ink}{RGB}{38,42,51}
\definecolor{stagefill}{RGB}{247,247,245}
\definecolor{stageline}{RGB}{120,124,130}
\definecolor{accent}{RGB}{31,106,110}
\definecolor{accentfill}{RGB}{233,242,242}
\definecolor{warm}{RGB}{178,82,52}

\newcommand{\code}[1]{\texttt{\small #1}}

\newcommand{\solkey}{page-level key}
\newcommand{\Solkey}{Page-level key}

\setlist{itemsep=1pt,topsep=2pt}

\newcommand{\papertitle}{Every Edge Needs a Quote: Building a Quote-Grounded Knowledge Graph of Healthcare Software Adoption with Language-Model Agents}
\hypersetup{pdftitle={\papertitle},pdfauthor={Juan Felipe Duran, Arthur Tobler, Martin Valli\`eres}}

\newcommand{\authorbox}[4]{\begin{tabular}[t]{@{}c@{}}\textbf{#1}\\#2\\#3\\\texttt{#4}\end{tabular}}

\begin{document}
\thispagestyle{firstpage}

\begin{center}
\rule{\linewidth}{4pt}\par\vspace{0.18in}
{\LARGE\bfseries \papertitle\par}
\vspace{0.18in}\rule{\linewidth}{1pt}\par\vspace{0.22in}
\authorbox{Juan Felipe Duran}{Medical Physics Unit, Department of Oncology}{McGill University, Montreal, Canada}{juan.duran@mail.mcgill.ca}%
\hfill
\authorbox{Arthur Tobler}{Toboggan Labs}{Montreal, Canada}{arthur@tobogganlabs.com}\\[0.16in]
\authorbox{Martin Valli\`eres}{Medical Physics Unit, Department of Oncology}{McGill University, Montreal, Canada}{martin.vallieres@mcgill.ca}
\par\vspace{0.1in}
{\small Correspondence: \texttt{juan.duran@mail.mcgill.ca}}
\end{center}
\vspace{0.1in}

\begin{abstract}
Hospitals regularly buy software that clinicians and patients depend on. Because these purchases are costly, hard to reverse and can affect patient care, buyers look for evidence of which peer institutions already use a product and which systems it connects to. This evidence is public but scattered across vendor websites, press releases and third-party pages. Language-model agents can now collect it at low cost, but the correctness of the extracted claims is difficult to establish, and a wrongly listed customer can mislead buyers and misrepresent an organisation. We present a pipeline that builds a knowledge graph of customer and integration relations from the open web under one rule: every edge must be supported by a verbatim quote from its source page. Pages come from a deterministic crawl of each vendor's site and a budgeted ReAct agent that searches the rest of the web. A permissive extractor proposes relations with quotes, and a layered grounding stage decides which to keep, using a referent step that reads the whole page when a quote names only one side, deterministic checks, and a panel of five model calls voting on nine criteria. On 61 vendors, the pipeline reaches 0.94 precision and 0.71 recall against a page-level key covering the 6,428 candidate relations proposed by its extractor, whose verdicts were drafted by one model and re-checked with a second. Against the golden reference, the marketplace's own curated list, it scores lower, but 89\% of the relations it keeps that the reference lacks are stated on a page. A confidence score built from the panel's votes keeps more correct relations than the fixed rule at the same precision, and reaches 0.90 precision at 0.90 recall. We also discuss how narrow, bounded model calls, string-matched evidence and voting address long-context degradation and run-to-run variance.
\par\vspace{0.6ex}\noindent\textbf{Keywords:} knowledge graphs, information extraction, language-model agents, attribution, LLM-as-a-judge, healthcare technology procurement.
\end{abstract}

% ===========================================================================
\section{Introduction}
\label{sec:intro}

Hospitals and health systems regularly need new software, from electronic health record modules and revenue-cycle tools to ambient scribes. These purchases are costly and hard to reverse once a product is built into clinical and administrative workflows, and they affect patients because clinicians use these tools while delivering care. Before buying, teams therefore look for evidence that a product works in practice: which peer institutions already use it, and which of their existing systems it connects to. This evidence is public, but it is spread across vendor case studies, testimonials, customer logos, press releases and integration marketplaces.

We developed our pipeline with Elion,\footnote{\url{https://elion.health/}} a marketplace that gathers this evidence so that buyers can compare products. For each product it records \emph{customers} (organisations that use the product) and \emph{integrations} (other products it connects to). In such a catalogue an incorrect customer is more harmful than a missing one: it can lead a buyer to trust a product for the wrong reason, and it associates a hospital with a product the hospital may not use.

Language models can now read web pages at scale and at low cost, which makes it possible to collect this evidence for every product in a catalogue. Their judgements, however, need checking. On one vendor's website, the same one-sentence testimonial, attributed to the same practice administrator, appears on several partner pages, each time under a different company's name. When we asked a large language model whether each of these pages shows that company using the product, it answered yes for three of them and no for a fourth, although the sentence was identical. More generally, models tend to say yes when asked whether a page supports a claim, can answer the same question differently on different runs, and produce worse output as the amount they must read or report at once increases.

The pipeline is built around one rule: \textbf{every edge needs a quote}. A relation is added to the graph only if it is supported by a verbatim span of a specific retained page, stored with the page's URL and retrieval time. This applies the notion of attribution of \citet{rashkin2023attribution} at the moment each edge is written. A quote lets anyone check an edge, but it does not make the edge correct, so most of the pipeline is concerned with deciding whether a quoted relation is actually supported.

We evaluate the pipeline on two benchmarks. The golden reference, the marketplace's own curated list, is accurate for the relations it lists but incomplete, so real relations it does not list count as errors. A page-level key labels every candidate relation. Its verdicts were drafted by one model and re-checked with a second.

\paragraph{Contributions.}
\begin{itemize}
  \item A quote-grounded pipeline that builds an adoption and integration graph from the open web and verifies every claim against the unmodified page (Sections~\ref{sec:task}--\ref{sec:system}).
  \item A page-level key of 6,428 candidate relations over 61 vendors, with every verdict re-checked (Section~\ref{sec:setup}).
  \item An evaluation on both benchmarks: 0.94 precision and 0.71 recall on the page-level key; and, against the golden reference, evidence that 89\% of the relations counted as errors are stated on a page (Section~\ref{sec:accuracy}).
  \item A confidence score built from the panel's votes that offers better operating points than the pipeline's fixed decision rule (Section~\ref{sec:score}).
  \item Design lessons on long-context degradation, run-to-run variance and prompt tuning (Section~\ref{sec:discussion}).
\end{itemize}

% ===========================================================================
\section{Problem Definition}
\label{sec:task}

The input is one target product $p$ and its vendor's website. The output is a set of typed edges
\begin{equation}
e = (s, r, o, q, u, \tau), \qquad r \in \{\textsc{Adopts}, \textsc{Integrates}\},
\label{eq:edge}
\end{equation}
where $s$ and $o$ are the subject and object entities, $q$ is the quote, $u$ is the URL of a retained page with text $T_u$, and $\tau$ is the retrieval time. For a customer relation (\textsc{Adopts}), $s$ is a health organisation (hospital, health system, clinic or practice) and $o = p$. For an integration relation (\textsc{Integrates}), $s = p$ and $o$ is another named product that is technically connected to it. The quote must occur in the normalised page text as one contiguous span, and each edge also records the identifier of the extraction run that produced it. A quote shows that a page stated a relation when it was retrieved; it does not establish that the relation is current, nor that the organisation endorses the product. Table~\ref{tab:terms} defines the terminology used in the remainder of the paper.

\begin{table}[h]
\centering
\small
\begin{tabularx}{\linewidth}{@{}lX@{}}
\toprule
Term & Meaning in this paper \\
\midrule
Customer relation & An organisation uses a product (\textsc{Adopts}). \\
Integration relation & Two named products are technically connected (\textsc{Integrates}). \\
Quote & The verbatim span of a page offered as evidence for a relation. \\
Candidate & One (page, relation) pair proposed by the extractor, with its quote. \\
Vendor page / third-party page & A page on the vendor's own site / on any other site. \\
Referent & The step that reads the whole page when a quote names only one side of a relation. \\
Semantic panel & Five independent model calls that vote on nine yes/no criteria. \\
Kept & Accepted by the pipeline and written to the graph. \\
Corpus pass & The single extraction run over every cached page of the 61 vendors that the evaluation scores. \\
Page-level key & Our benchmark: 6,428 candidates, each labelled ``does this page support this relation?'' \\
Golden reference & The marketplace's own list of customers and integrations for each vendor. \\
\bottomrule
\end{tabularx}
\caption{Terminology used throughout the paper.}
\label{tab:terms}
\end{table}

% ===========================================================================
\section{Method}
\label{sec:system}

Figure~\ref{fig:pipeline} shows the pipeline, and Figure~\ref{fig:example} follows one real relation through it. The pipeline has three stages: discovery finds pages, extraction proposes relations, and grounding decides which relations to keep. It is orchestrated with LangGraph~\citep{langgraph}, model calls go through OpenRouter~\citep{openrouter}, and every prompt is versioned in Braintrust~\citep{braintrust}. Extraction reads only an immutable cached copy of each page, never the live web. Appendix~\ref{app:impl} gives implementation details.

\begin{figure}[htbp]
\centering
\resizebox{\linewidth}{!}{%
\begin{tikzpicture}[
  font=\small,
  box/.style={draw=stageline, rounded corners=3pt, align=center, fill=stagefill,
              line width=0.6pt, text=ink, inner sep=3pt, minimum height=9mm},
  llm/.style={draw=accent, rounded corners=3pt, align=center, fill=accentfill,
              line width=1pt, text=ink, inner sep=3pt, minimum height=9mm},
  arr/.style={-{Stealth[length=2.2mm]}, line width=0.8pt, ink!75},
  grp/.style={draw=black!55, rounded corners=4pt, line width=0.7pt, inner sep=5pt},
  lab/.style={font=\footnotesize\bfseries, text=ink}
]
% Discovery: first party
\node[box, text width=19mm] (map) {Map site\\{\scriptsize Firecrawl, $\le$1,000 URLs}};
\node[box, text width=19mm, right=4mm of map] (fetch) {Fetch \& cache\\{\scriptsize immutable copy}};
\node[box, text width=19mm, right=4mm of fetch] (gate) {Relevance gate\\{\scriptsize names the product?}};
% Discovery: third party
\node[llm, text width=19mm, below=14mm of map] (search) {Search\\{\scriptsize agent writes queries}};
\node[llm, text width=19mm, right=4mm of search] (read) {Read page\\{\scriptsize only surfaced URLs}};
\node[llm, text width=19mm, right=4mm of read] (keep) {Keep?\\{\scriptsize agent's decision}};
\draw[arr] (map) -- (fetch); \draw[arr] (fetch) -- (gate);
\draw[arr] (search) -- (read); \draw[arr] (read) -- (keep);
\draw[arr, accent] (keep.south) to[bend left=22] (search.south);
\coordinate (loopbot) at ($(read.south)+(0,-4.5mm)$);
\draw[arr, dashed, warm] (gate.south) -- node[left, font=\scriptsize, text=warm, align=right]{$\le$8 customer names\\as search leads} (keep.north);
\begin{scope}[on background layer]
\node[grp, fit=(map)(fetch)(gate), label={[lab]above:{Vendor's own site (deterministic)}}] (fp) {};
\node[grp, fit=(search)(read)(keep)(loopbot), label={[lab]below:{Rest of the web (ReAct agent, loop under a tool budget)}}] (tp) {};
\end{scope}

% Extraction
\node[llm, text width=21mm, right=13mm of gate, yshift=-10mm] (ext) {\textbf{Extractor}\\{\scriptsize relations + quote,\\ permissive}};
\coordinate (join) at ($(ext.west)+(-5mm,0)$);
\draw[arr] (fp.east) -| (join) -- (ext.west);
\draw[arr] (tp.east) -| (join) -- (ext.west);

% Grounding
\node[box, text width=16mm, right=9mm of ext] (splice) {Splice\\repair};
\node[llm, text width=18mm, right=4mm of splice] (ref) {Referent\\{\scriptsize reads whole page}};
\node[box, text width=16mm, right=4mm of ref] (shape) {Shape\\checks {\scriptsize (14)}};
\node[llm, text width=19mm, right=4mm of shape] (sem) {Semantic panel\\{\scriptsize 5 votes $\times$ 9 criteria}};
\draw[arr] (ext) -- (splice); \draw[arr] (splice) -- (ref); \draw[arr] (ref) -- (shape); \draw[arr] (shape) -- (sem);
\begin{scope}[on background layer]
\node[grp, fit=(splice)(ref)(shape)(sem), label={[lab]above:{Grounding (against the unmodified page)}}] (gr) {};
\end{scope}

% Graph
\node[box, text width=20mm, right=9mm of sem, minimum height=14mm] (neo) {\textbf{Neo4j graph}\\{\scriptsize edge + quote,\\ URL, time, run id}};
\draw[arr] (gr.east) -- (neo.west);

% Legend
\node[llm, minimum height=4mm, minimum width=6mm, inner sep=1pt, below=11mm of shape, xshift=-14mm] (lg1) {};
\node[right=1mm of lg1, font=\scriptsize, text=ink] {model call};
\node[box, minimum height=4mm, minimum width=6mm, inner sep=1pt, right=16mm of lg1] (lg2) {};
\node[right=1mm of lg2, font=\scriptsize, text=ink] {deterministic};
\end{tikzpicture}%
}
\caption{The pipeline. Discovery finds pages in two halves: a deterministic sweep of the vendor's own site, and a ReAct agent that searches everything else under a tool budget. Customer names confirmed on the vendor's site become search leads for the agent. Both halves feed the same extractor and the same grounding stage, which run in the order shown. Only relations that survive every step are written, each with its quote and provenance.}
\label{fig:pipeline}
\end{figure}
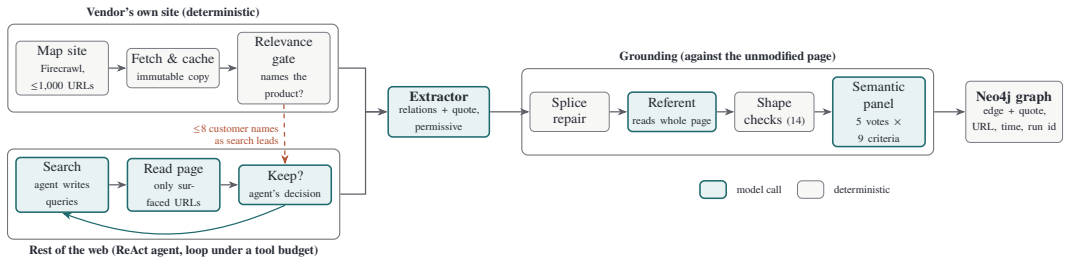
\begin{figure}[htbp]
\centering
\fbox{\begin{minipage}{0.96\linewidth}
\small
\textbf{Worked example: one relation through the pipeline}\par\smallskip
\begin{enumerate}[leftmargin=1.4em, itemsep=3pt]
  \item \textbf{Discovery.} The crawl of Abridge's own site retrieves a blog post titled \emph{Inside Northwestern Medicine's Evidence-Based Evaluation of AI}. It names the product, so it passes the relevance gate and is cached.
  \item \textbf{Extraction.} The extractor proposes \emph{Northwestern Medicine} \textsc{Adopts} \emph{Abridge}, with the quote: ``Today, after 3 months, we're already reaching the point where clinician demand exceeds our available licenses.''
  \item \textbf{Referent.} The quote names neither the hospital nor the product, so the referent reads the whole page with the prompt for customer relations on vendor pages. It confirms, arguing that ``the Abridge-owned page describes Northwestern Medicine selecting Abridge after a multi-vendor evaluation and reports 1,300+ clinicians live with active utilization, establishing current use by the named health system.''
  \item \textbf{Shape checks.} The subject is a named organisation and the quote is on the page as one span, so nothing is rejected.
  \item \textbf{Semantic panel.} All five calls say yes on each of the five criteria that gate a customer relation. One call's reason notes that the testimonial's ``we'' refers to Northwestern Medicine, which is the reference the referent step is designed to resolve.
  \item \textbf{Graph.} The relation is kept; in the live pipeline it is written to Neo4j with its quote, the page URL and the retrieval time. The page-level key agrees that the page supports it.
\end{enumerate}
\end{minipage}}
\caption{A relation from the evaluation corpus, with the referent's argument and the panel's votes as recorded. Section~\ref{sec:grounding} describes each grounding step in detail.}
\label{fig:example}
\end{figure}

\subsection{Discovery}
\paragraph{Vendor pages.}
Because the vendor's domain is known, this half is a deterministic sweep with no model calls. A Firecrawl~\citep{firecrawl} map call lists up to 1,000 URLs from the site root, with the product's own section of the site listed first. URLs are normalised, technical pages (static assets, login pages, sitemaps, developer documentation) are excluded with a recorded reason, and the remaining pages are scraped as Markdown and cached with their retrieval time. Some evidence only appears after interaction, so for each vendor page a headless browser also opens up to five tabs, filters or ``load more'' buttons, and records the alternative text of logos placed under evidence headings. Finally, a relevance gate drops any page that does not name the product; dropped pages are kept on record but never extracted.

\paragraph{Third-party pages.}
Outside the vendor's site, the set of relevant pages is not known in advance, so an agent searches for them. One ReAct-style agent~\citep{yao2023react} per relation type can search the web, map and crawl sites, and read pages. Every tool call must state a reason, which is logged, so each decision can be audited. The agent cannot read the vendor's own domain, and it cannot read a short list of hosts that only add noise: social networks, job boards, app stores, the marketplace's own site, and three aggregator sites that produced fabricated relations. It can only read a page whose URL a search, map or crawl has already returned.

The agent works under hard budgets (25 searches and 15 page reads per agent), and each page it reads is truncated to 20,000 characters. Before it starts, it receives up to eight customer names already confirmed on vendor pages as search leads. It ends by returning the pages it wants to keep, each with a reason. Those pages are cached and pass through exactly the same extraction and grounding as vendor pages.

\subsection{Extraction}
Extraction is one structured-output call per (page, relation type) over the whole page, with no chunking. The prompt receives the target product's name and the graph entities already known to appear on the page, and returns a list of relations, each with a quote. The extractor is deliberately permissive: when the wording is ambiguous it extracts anyway, and leaves the decision to grounding. Three instructions carry most of the weight:
\begin{itemize}
  \item quote \emph{one contiguous evidence span, copied from the visible text}, so that every quote can be verified;
  \item \emph{include enough adjacent heading or sentences for both names and the relationship cue to appear in that span}, because proof is often split between a heading and the line under it;
  \item adoption is \emph{not an investment, an award, or network membership}, stated as explicitly as what does count.
\end{itemize}
Before grounding, each endpoint is matched to the entities already in the graph, so that one hospital under several spellings is not counted several times. A name that exactly matches a known entity on the page is resolved directly; otherwise Graphiti's resolver~\citep{rasmussen2025zep} tries normalised names, fuzzy matching and finally a model call. Only health organisations may be added as new entities; a product that matches nothing is rejected.

\subsection{Grounding}
\label{sec:grounding}
Grounding applies the following four steps in order (Algorithm~\ref{alg:grounding}). Every step reads the original page text, never a modified copy.

\begin{algorithm}[tbp]
\caption{Grounding one candidate relation}
\label{alg:grounding}
\small
\begin{algorithmic}[1]
\Require unmodified page text $T$, candidate $c = (s, r, o, q)$
\Ensure keep $c$, or reject it with a reason code
\State $\mathit{spliced} \gets \textbf{false}$
\If{$q$ matches $T$ (at least 75\% of its characters) but not as one contiguous span}
  \State $q \gets$ the span of $T$ covering the matched pieces \Comment{splice repair}
  \State $\mathit{spliced} \gets \textbf{true}$
\EndIf
\If{$q$ does not name both $s$ and $o$, \textbf{or} $\mathit{spliced}$}
  \State $(\mathit{argument}, \mathit{verdict}, \mathit{proof}) \gets \textsc{Referent}(T, s, r, o)$ \Comment{reads the whole page}
  \If{$\mathit{verdict}$ is no \textbf{or} $\mathit{proof}$ is not found in $T$}
    \State \Return reject (referent declined)
  \EndIf
\EndIf
\For{each shape check $g$}
  \If{$g(c, T)$ fails} \State \Return reject (code of $g$) \EndIf
\EndFor
\State $E \gets$ section of $T$ that best matches $q$ (at most 6,000 characters), plus $\mathit{proof}$ and $\mathit{argument}$ if available
\State run five panel calls on $(c, E)$ and count the votes on every criterion
\If{a criterion that applies to $r$ gets fewer than half yes votes}
  \State \Return reject (first failing criterion)
\EndIf
\State \Return keep, and write $(s, r, o, q, u, \tau)$ to the graph
\end{algorithmic}
\end{algorithm}

\paragraph{Splice repair.}
Asked for one unbroken span, the extractor sometimes joins two genuine pieces of the page and drops the few characters between them, for example between a testimonial and its attribution line. This happens to about one quote in eleven. When the quote matches the page (at least 75\% of its characters) but is not one contiguous run, it is replaced by the page's own text spanning the matched pieces, and the relation is sent to the referent.

\paragraph{Referent.}
Many quotes name only one side of a relation: a testimonial names the hospital but not the product, or the product appears only in the page title. The referent (in the linguistic sense: the thing a word such as \emph{we} refers to) runs when the quote does not name both sides, or when the quote was repaired. It reads the whole page and returns an argument, a yes/no verdict and a proof sentence copied from the page, which must be found in the original text. A refusal rejects the relation. There are four referent prompts, one per relation type and page origin. On a vendor's own site, a case study or customer spotlight is the vendor presenting a customer; on a third party's page, a list of names often describes the page owner's own clients (Appendix~\ref{app:prompts}). The referent is not shown the extractor's quote. When it confirms, its proof sentence and argument are passed to the semantic panel, labelled as a claim from an earlier step.

\paragraph{Shape checks.}
Fourteen deterministic checks (Appendix~\ref{app:checks}) reject structural failures at no model cost, such as a generic endpoint (``hospitals''), a person in place of an organisation, a quote that is not on the page, a self-edge, or a name that appears only inside a link's URL.

\paragraph{Semantic panel.}
Five identical model calls judge each remaining relation independently. Each call answers nine yes/no criteria (Table~\ref{tab:criteria}), and a criterion passes when at least half of the returned votes are yes. A relation is rejected if any criterion that applies to its type fails: five criteria apply to customer relations and seven to integrations. The panel does not see the whole page: it sees the section that best matches the quote, found by splitting the page at its headings and capped at 6,000 characters, plus anything the referent passed on. All five votes are recorded for every relation.

\subsection{Persistence}
Surviving relations are written to a Neo4j graph~\citep{neo4j} through Graphiti. Organisations and products are nodes, and each relation is an edge carrying its quote, source URL, retrieval time, extraction identifier and the version of the prompts that produced it. Every rejection is stored with its reason, which makes the analysis in Section~\ref{sec:losses} possible without re-running anything.

% ===========================================================================
\section{Experimental Setup}
\label{sec:setup}

We use one corpus and two benchmarks that answer different questions: the page-level key asks whether a page states a relation, and the golden reference is the marketplace's own list of customers and integrations.

\paragraph{Corpus.}
The evaluation covers the 61 products in the golden reference. We ran extraction once over every cached page for these products (9 October 2026), recording for every candidate whether it was kept, the gate that rejected it, the referent's decision and each of the five votes. All model calls in this pass used \code{openai/gpt-6-luna}, with every prompt pinned to a recorded version. Two grounding components were finalised after this pass: the splice repair and the referent prompt for customer relations on vendor pages. The candidates they affect (570 with spliced quotes and 306 that the referent had refused) were re-run through the final grounding stage, so the results describe the final pipeline. The pass produced 6,429 candidate relations from 3,284 pages: 2,952 on vendor sites and 332 third-party pages found by the discovery agent. One candidate could not be judged, because its organisation is named only in an image's alt text, which leaves 6,428 (Table~\ref{tab:corpus}).

\begin{table}[htbp]
\centering
\small
\begin{tabular}{lrrr}
\toprule
& Customers & Integrations & Total \\
\midrule
\textbf{Page-level key}: relations proposed by the extractor & 3,486 & 2,942 & 6,428 \\
\quad of which correct & 2,952 & 1,861 & 4,813 \\
\midrule
\textbf{Golden reference}: relations listed & 749 & 249 & 998 \\
\bottomrule
\end{tabular}
\caption{The two benchmarks over the 61 vendors.}
\label{tab:corpus}
\end{table}

\paragraph{Golden reference.}
The golden reference is the marketplace's own record, exported from its production database and maintained by its annotation team: 749 customers and 249 integrations over the 61 products. Four products list no customers and fifteen list no integrations. Its unit is the pair (vendor, organisation), so page-level output is collapsed before scoring: a pair counts as predicted if any page kept it. Names are matched after normalisation with a string-similarity threshold of 0.82, which absorbs spelling variants. The reference is correct about what it lists but incomplete in two ways. It lists organisations that no page in the corpus mentions, so recall measured against it is a lower bound. And many real customers that vendors publish are missing from it, so precision measured against it is too low. For 239 customers and 51 integrations, the only proof is on third-party pages, which only the discovery agent can reach.

\paragraph{Page-level key.}
The page-level key asks: ``does \emph{this page} support \emph{this} relation?'' Every candidate, kept or not, was judged by \code{openai/gpt-5.6-sol} (GPT-5.6 Sol), a stronger model tier than the pipeline's. It read the whole page and answered YES or NO with a one-sentence justification (prompts in Appendix~\ref{app:prompts}). There is one verdict per distinct (page, relation), so the same relation on two pages is judged twice. Because every candidate is labelled, the decision threshold can be moved in both directions (Section~\ref{sec:score}). Claude Opus 5 (Anthropic) then re-checked every verdict against about 900 characters of page text around the quote, and the first author read a subset (the records do not mark which entries). Of the 6,428 verdicts, 911 (14\%) were reversed, each with a written reason. A candidate counts as correct if its final verdict is YES.

\paragraph{Metrics.}
With true positives (TP), false positives (FP) and false negatives (FN) counted at each benchmark's own unit, we report precision $P = \mathrm{TP}/(\mathrm{TP}+\mathrm{FP})$, recall $R = \mathrm{TP}/(\mathrm{TP}+\mathrm{FN})$ and $F_1 = 2PR/(P+R)$. On the page-level key, recall is measured over the candidates the extractor proposed, so it describes the grounding stages; a relation the extractor never proposed is invisible to this key. The golden reference does see extraction recall, but mixes it with its own gaps.

% ===========================================================================
\section{Results}
\label{sec:results}

We report, in order, the pipeline's accuracy on both benchmarks, where it loses correct relations, what the referent contributes, and what a confidence score adds.

\subsection{Overall Accuracy}
\label{sec:accuracy}

\begin{table}[htbp]
\centering
\small
\begin{tabular}{llrrr}
\toprule
Benchmark & Scope & Precision & Recall & F1 \\
\midrule
\multirow{6}{*}{\Solkey{} (page, relation)}
 & All & \textbf{0.94} & \textbf{0.71} & \textbf{0.81} \\
 & \quad by type: customers & 0.96 & 0.71 & 0.81 \\
 & \quad by type: integrations & 0.90 & 0.71 & 0.79 \\
 & \quad by page: vendor's site & 0.94 & 0.71 & 0.81 \\
 & \quad by page: third-party & 0.92 & 0.65 & 0.76 \\
 & Each relation counted once (2,349) & 0.92 & 0.74 & 0.82 \\
\midrule
\multirow{3}{*}{Golden reference}
 & All & 0.34 & 0.41 & 0.37 \\
 & \quad by type: customers & 0.43 & 0.42 & 0.42 \\
 & \quad by type: integrations & 0.20 & 0.39 & 0.27 \\
\bottomrule
\end{tabular}
\caption{Accuracy on the 61-vendor corpus. On the page-level key, recall is over the candidates the extractor proposed; on the golden reference, over every organisation the reference lists. ``Each relation counted once'' counts a relation as kept if any page kept it and as correct if any page states it. The ``by type'' and ``by page'' rows are separate splits of the same candidates.}
\label{tab:accuracy}
\end{table}

On the page-level key, 94\% of the relations the pipeline keeps are correct, and the pipeline keeps 71\% of the correct candidates (Table~\ref{tab:accuracy}). Precision is high for both relation types, and recall is the same for both. Third-party pages are judged nearly as precisely as vendor pages, with lower recall. Vendor sites repeat some claims on many pages (one sentence about Sutter Health using Abridge appears on 179 pages), but counting each unique relation once gives the same picture (0.92 / 0.74 / 0.82).

Against the golden reference the same output scores 0.34 precision and 0.41 recall, and both numbers mostly reflect the reference. On recall, the reference lists organisations that no page in the corpus mentions (Section~\ref{sec:setup}). On precision, we checked every vendor--organisation pair the pipeline keeps that the reference does not list. Of these 803 pairs, 716 (89\%) rest on at least one page that the page-level key marks as stating the relation (388 of 420 customer pairs and 328 of 383 integration pairs). Most of them are real relations the reference does not yet include. Integration precision is lowest because the reference lists only 249 integrations in total, and none for fifteen of the products.

The discovery agent was run once on all 61 vendors for customer relations (US\$2.19 in total). Of the 239 customers whose only proof is on third-party pages (Section~\ref{sec:setup}), it found pages proving 88 (37\%).

\subsection{Error Analysis}
\label{sec:losses}

\begin{table}[htbp]
\centering
\small
\begin{tabular}{lrr}
\toprule
Outcome & Correct & Incorrect \\
\midrule
Kept & 3,415 & 217 \\
Rejected by the semantic panel & 1,106 & 1,212 \\
\quad page does not present them as users & 546 & 305 \\
\quad all other criteria & 560 & 907 \\
Refused by the referent & 276 & 108 \\
Rejected by the shape checks & 16 & 78 \\
\midrule
Total & 4,813 & 1,615 \\
\bottomrule
\end{tabular}
\caption{Outcome of the 6,428 judged candidates under the final pipeline, by the step that rejected them. Correct and incorrect are according to the page-level key.}
\label{tab:losses}
\end{table}

Table~\ref{tab:losses} traces every candidate to the step that rejected it. Of the 1,398 correct relations the pipeline misses, the semantic panel rejects 1,106 and the referent 276. The deterministic checks lose only 16 correct relations and block 78 incorrect ones. Within the panel, the criterion that rejects the most correct relations is whether the page presents the organisation as a user of the product (546 correct, 305 incorrect), which often depends on context outside the quote. Of the relations the panel rejects, 594 had already been confirmed by the referent after reading the whole page, and 391 of these are correct. The panel reads at most 6,000 characters around the quote, so it misses evidence elsewhere on the page, such as a ``Trusted by'' heading above a row of logos or the attribution under a testimonial. Section~\ref{sec:fatigue} discusses this trade-off.

\subsection{Effect of the Referent}
\label{sec:components}

\begin{table}[htbp]
\centering
\small
\begin{tabular}{lrrrrr}
\toprule
Configuration & Kept & P & R & F1 & Golden F1 \\
\midrule
Without the referent & 133 & 0.887 & 0.361 & 0.513 & 0.578 \\
Final pipeline & 291 & 0.932 & 0.808 & 0.866 & 0.585 \\
\bottomrule
\end{tabular}
\caption{Customer relations on 266 vendor pages from 21 vendors, run end to end (one run each). P/R/F1 are against this set's reviewed verdicts; golden F1 counts only customers that appear on these pages.}
\label{tab:referent}
\end{table}

This comparison uses a separate, smaller set: customer relations on 266 vendor pages from 21 vendors, scored against their own reviewed verdicts. Its numbers are not directly comparable with Table~\ref{tab:accuracy}. Customer evidence on a vendor's site is often split across a page: a testimonial names the hospital, while the title or a heading names the product. On such pages (Table~\ref{tab:referent}), the referent more than doubles the number of relations kept (from 133 to 291), raising recall from 0.361 to 0.808 and precision from 0.887 to 0.932. The golden reference barely changes (F1 0.578 against 0.585), because most of the relations the referent recovers have no golden row. Only the page-level key detects this gain.

\subsection{Confidence Score}
\label{sec:score}

\begin{figure}[htbp]
\centering
\begin{tikzpicture}
\begin{axis}[
  width=0.78\linewidth, height=0.46\linewidth,
  xlabel={Recall (\solkey{})}, ylabel={Precision (\solkey{})},
  xmin=0, xmax=1.0, ymin=0.72, ymax=1.0,
  xtick={0,0.2,0.4,0.6,0.8,1.0}, ytick={0.75,0.80,0.85,0.90,0.95,1.00},
  yticklabel style={/pgf/number format/fixed,/pgf/number format/fixed zerofill,/pgf/number format/precision=2},
  grid=major, grid style={black!10},
  tick label style={font=\footnotesize}, label style={font=\small},
  legend style={font=\footnotesize, at={(0.02,0.04)}, anchor=south west, draw=black!20},
]
\addplot[very thick, color=accent] table[x=recall, y=precision] {data/pr_curve.dat};
\addlegendentry{confidence score, threshold swept}
\addplot[only marks, mark=*, mark size=2.6pt, color=warm] coordinates {(0.63,0.94)};
\addlegendentry{pipeline's fixed decision}
\addplot[only marks, mark=o, mark size=2pt, color=accent] coordinates {(0.561,0.956) (0.745,0.946) (0.816,0.936) (0.875,0.912)};
\node[font=\scriptsize, anchor=south] at (axis cs:0.561,0.958) {0.90};
\node[font=\scriptsize, anchor=south] at (axis cs:0.745,0.948) {0.85};
\node[font=\scriptsize, anchor=south west] at (axis cs:0.816,0.938) {0.80};
\node[font=\scriptsize, anchor=south west] at (axis cs:0.875,0.914) {0.70};
\end{axis}
\end{tikzpicture}
\caption{Precision--recall curve of the confidence score, with every relation scored by a model that did not see it in training (five-fold cross-validation, all 22 features). Open circles mark score thresholds. The filled dot is the pipeline's fixed decision on the same run (precision 0.94, recall 0.63). At that precision, the score reaches a recall of about 0.77.}
\label{fig:pr}
\end{figure}
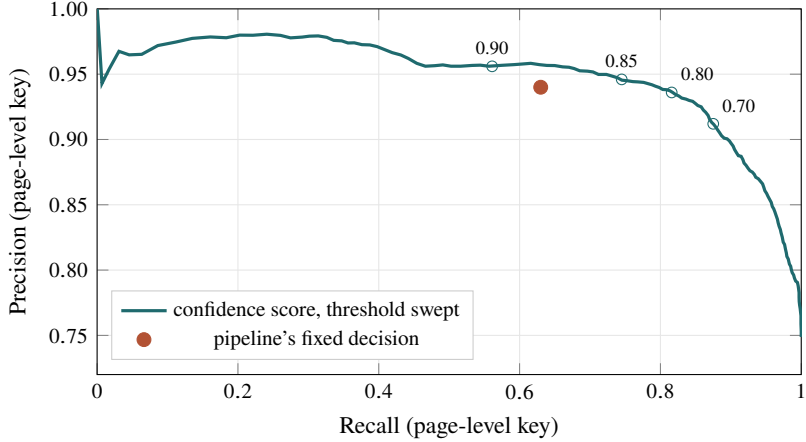

The pipeline's fixed decision rule gives one point on the precision--recall trade-off. To offer other points, we fit a score offline from the signals recorded in the corpus pass, with no further model calls. Each relation $c$ is described by a vector $\mathbf{x}_c \in \mathbb{R}^{22}$ of recorded signals: eight are the share of the five panel calls that said yes on a criterion, eleven are flags describing how the relation was handled (for example, whether the referent was asked and whether it confirmed), and three are counts (the quote length, how often the quote is reused on the page, and how many relations the extractor emitted in the same call). We fitted a logistic regression,
\begin{equation}
\mathrm{score}(c) = \sigma\big(\mathbf{w}^{\top}\mathbf{x}_c + b\big), \qquad \mathrm{keep}_\theta(c) = \mathbf{1}\big[\mathrm{score}(c) \geq \theta\big],
\label{eq:score}
\end{equation}
where $\sigma$ is the logistic function and $\theta$ is a threshold chosen by the user. We evaluate it out-of-sample with five-fold cross-validation against the page-level key. To see which signals matter, we also searched feature subsets and shallow decision trees (Appendix~\ref{app:score}).

The score reaches an AUC of 0.887. Figure~\ref{fig:pr} and Table~\ref{tab:operating} show the trade-off it exposes: at 0.90 precision it keeps 0.90 of the correct relations, at 0.95 precision it keeps 0.72, and at the pipeline's own precision of 0.94 it keeps about 0.77, against 0.63 for the fixed rule on the same run. (Table~\ref{tab:accuracy} reports 0.71 because it includes the re-run described in Section~\ref{sec:setup}.) A precision of 0.99 is reached only for the twelve highest-scoring relations; beyond that point, relations with identical signals can differ in whether they are correct. The catalogue's owners can choose the threshold and send relations below it to human review.

\begin{table}[htbp]
\centering
\small
\begin{tabular}{rrr}
\toprule
Target precision & Recall & Relations kept \\
\midrule
0.99 & 0.002 & 12 \\
0.98 & 0.27 & 1,305 \\
0.97 & 0.40 & 2,002 \\
0.96 & 0.45 & 2,258 \\
0.95 & 0.72 & 3,623 \\
0.93 & 0.84 & 4,359 \\
0.90 & 0.90 & 4,792 \\
\midrule
\multicolumn{3}{l}{Pipeline's fixed decision, same run: precision 0.94, recall 0.63} \\
\bottomrule
\end{tabular}
\caption{Operating points of the confidence score on the page-level key, out-of-sample.}
\label{tab:operating}
\end{table}

 Most of the signal comes from the panel's votes. The share of the five calls agreeing that the claim is supported reaches F1 0.864 on its own, and a model with five features performs nearly as well as the full model (Appendix~\ref{app:score}). The referent helps when the panel is split: relations the referent confirmed are then 0.89 correct, and those it refused 0.33. Two cautions apply. First, the heaviest weight in most fitted models is the flag marking a customer relation, which reflects this corpus's base rate rather than evidence about the relation, and should be removed before the score is used elsewhere. Second, no signal describes what kind of page the evidence came from (a case study, a logo wall, a navigation banner repeated across the site), and that is where the remaining errors concentrate.

% ===========================================================================
\section{Discussion}
\label{sec:discussion}

Two properties of language models shaped the design more than the choice of model: their output degrades as their input or output grows, and it varies from run to run. This section describes how the pipeline handles each and how its prompts were tuned.

\subsection{Long-Context Degradation}
\label{sec:fatigue}

A model's output degrades as the amount it must read or report at once grows~\citep{liu2024lost}, and the corpus shows this directly. The extractor's splicing rate rises with the number of relations it must report in one call: 6 in 100 quotes when it reports one relation, and 15 in 100 when it reports four to seven. The discovery agent faces the same pressure, because search results and page text accumulate over a session. The pipeline responds in three ways.
\begin{itemize}
  \item \emph{Narrow, bounded calls.} Each model call has one task and a limited input. The referent answers one question about one relation; the panel answers nine fixed criteria on at most 6,000 characters; the agent reads pages truncated to 20,000 characters, under a hard budget.
  \item \emph{Deterministic checks wherever possible.} No model is asked whether a quote is on the page; quotes and proof sentences are matched by string comparison, and spliced quotes are repaired deterministically.
  \item \emph{Reading separated from judging.} The referent is not shown the extractor's quote. On a 216-case set, the referent accepted 162 of 176 true relations without the quote and 141 with it, which suggests that a weak quote anchors the model.
\end{itemize}
Bounded inputs also have a cost: most correct relations that the panel rejects after the referent confirmed them have their evidence outside the panel's excerpt (Section~\ref{sec:losses}). Widening the panel's view, or letting a referent confirmation count as evidence, is the most direct fix.

\subsection{Run-to-Run Variance}
The same input can receive different answers on different runs: three identical extraction calls on one page returned 1, 39 and 2 relations. The pipeline uses five calls per relation with a vote per criterion, and records every vote. The vote shares are the strongest features of the confidence score (Section~\ref{sec:score}). Prompt comparisons use several runs per version on frozen inputs (Section~\ref{sec:tuning}).

\subsection{Prompt Tuning}
\label{sec:tuning}
All prompts are versioned, and changing any grounding prompt re-runs extraction under a new version. Prompts were tuned with one protocol:
\begin{enumerate}
  \item \emph{Freeze the inputs.} Each case set is a fixed set of pages with fixed candidate relations, so differences between versions come from the prompt and not from extraction.
  \item \emph{Label and review.} The judge model labels each case, and its verdicts are corrected where review finds them wrong.
  \item \emph{Repeat and pair.} Competing versions run several times on the same cases (up to seven runs per finalist) and are compared case by case with paired tests.
  \item \emph{Confirm end to end.} The chosen version is run through the whole pipeline, because a later stage can absorb or undo what a step-level score rewards.
\end{enumerate}
The prompts share a few principles. The semantic panel asks for explicit criteria as separate yes/no fields rather than one overall verdict. The grounding prompts state what does not count (an investment, an award, network membership, a bare logo) as clearly as what does. The referent prompts give a worked YES example and a worked NO example, ask for the argument before the verdict, and ask for a proof sentence copied verbatim so that it can be string-matched. Separate referent prompts for vendor and third-party pages also help: on 216 customer relations on vendor pages, the vendor-page prompt reaches F1 0.975 over seven runs, against 0.955 for a single generic prompt over three (Wilcoxon signed-rank $p = 0.014$).

% ===========================================================================
\section{Related Work}
\label{sec:related}

\paragraph{Knowledge-graph construction.}
Web-scale knowledge bases such as Knowledge Vault fuse noisy extractions into probabilistic facts~\citep{dong2014knowledgevault}. Distant supervision uses an existing knowledge base as labels and inherits its gaps as false negatives~\citep{mintz2009distant}, a positive--unlabelled setting~\citep{elkan2008pu} that our golden reference shares. Language models are now widely used for relation extraction and graph construction~\citep{wadhwa2023revisiting,pan2024unifying,bian2025kgsurvey}; GraphJudge uses a model to filter candidate triples~\citep{huang2025graphjudge}, close in spirit to our semantic panel. We store the graph with Graphiti, the engine behind Zep's temporal knowledge graph~\citep{rasmussen2025zep}.

\paragraph{Attribution and verification.}
Attribution to identified sources formalises when a statement is supported by a source~\citep{rashkin2023attribution}. Related work asks models to cite~\citep{gao2023alce}, revises output against retrieved evidence~\citep{gao2023rarr}, and scores factual precision of atomic claims~\citep{min2023factscore}; \citet{ji2023hallucination} survey unsupported generation. We apply attribution when each edge is written and verify against the unmodified page.

\paragraph{Agents and model judges.}
Our discovery agent follows ReAct~\citep{yao2023react}, as browsing agents such as WebGPT do~\citep{nakano2021webgpt}; retrieval-augmented generation is the broader family~\citep{lewis2020rag}. Models used as judges are useful but biased and inconsistent~\citep{zheng2023judging}, and majority voting over samples improves reliability~\citep{wang2023selfconsistency}; our panel votes per criterion and records every vote.

\paragraph{Quality of evaluation data.}
Label errors in test sets can change which model appears best~\citep{northcutt2021pervasive}. Every verdict of our page-level key was re-checked. Treating reference data as a component to audit, rather than fixed infrastructure, follows the internal-audit framing of \citet{raji2020closing} and the system-level view of \citet{sculley2015debt}. Our confidence score is evaluated by cross-validation, and calibration in the sense of \citet{guo2017calibration} is a natural next step.

% ===========================================================================
\section{Limitations and Responsible Use}
\label{sec:limits}

\paragraph{Benchmarks.} The page-level key covers only relations the extractor proposed, so extraction recall is visible only through the golden reference, which mixes it with the reference's own gaps. The key was re-checked by a second model and only partly by a person, with no measured agreement rate. The review saw the judge's verdicts, which may anchor it toward agreement, and the pipeline and judge models come from the same provider. A human-annotated sample with measured agreement would bound the key's error rate. The golden reference is maintained by the industry partner for whom the system was built, all vendors come from its catalogue, and we have not evaluated other domains.

\paragraph{Measurement.} The final scores combine the corpus pass with a re-run of the candidates affected by the final splice repair and referent prompt, rather than one fresh pass. The confidence score was fitted on signals from the corpus pass, on one corpus, and several component measurements rest on one to seven runs.

\paragraph{What a quote establishes.} A quote shows that a page stated a relation when it was retrieved. It does not show that the relation is current, and a vendor's public claim does not show that an organisation endorses the product or agreed to be named. The graph does not yet model time beyond retrieval dates. Logo walls rendered as images are only partly readable, through their alternative text. Entity resolution is incomplete, so the same hospital can still appear under several names. Records should be reviewed by a person before they inform a purchasing decision, and no threshold for publishing without review has been validated.

% ===========================================================================
\section{Conclusion}
\label{sec:conclusion}

Language-model agents make it inexpensive to extract claims from the web, but those claims must be verified before anyone acts on them. We presented a pipeline that builds a graph of healthcare software adoption and integration in which every edge points to its source sentence and page, and in which narrow model calls and deterministic checks decide which quoted relations to keep. On 61 vendors it reaches 0.94 precision and 0.71 recall against a re-checked page-level key, and most of the relations it finds beyond the golden reference are stated on a page. A confidence score built from its recorded votes lets the catalogue's owners choose their own balance between precision and recall. Our design recommendations are: give each model call one task and a bounded input, check deterministically whatever can be checked, use votes instead of single calls, and review the evaluation labels before relying on scores.

\section*{Code Availability}
The code is available at \url{https://github.com/MEDomicsLab/healthcare-agentic-kg}. The repository contains the full pipeline (vendor-site crawl, discovery agent, extraction, grounding and persistence), public adaptations of its 20 prompts, ten illustrative products with 139 customer and 21 integration labels, and public editions of the experiment write-ups. Its test suite runs without API keys or network access, and the pipeline can be run on the ten products with one's own keys for page fetching and model calls. Because the prompts are adapted, the sample is small and web pages change, a re-run will not reproduce the full-corpus results reported here; the artefacts behind those results are listed in Appendix~\ref{app:repro}.

\section*{Acknowledgments}
This work was supported by Mitacs, Mitacs Accelerate program (\#FR170683).

\section*{Use of AI Assistants}
Claude Opus 5 (Anthropic) was used to re-check the page-level key (Section~\ref{sec:setup}), to help write the pipeline's code and to help draft this manuscript. The authors take full responsibility for its content.

\bibliographystyle{plainnat}
\bibliography{references}

\begin{thebibliography}{28}
\providecommand{\natexlab}[1]{#1}
\providecommand{\url}[1]{\texttt{#1}}
\expandafter\ifx\csname urlstyle\endcsname\relax
  \providecommand{\doi}[1]{doi: #1}\else
  \providecommand{\doi}{doi: \begingroup \urlstyle{rm}\Url}\fi

\bibitem[Bian(2025)]{bian2025kgsurvey}
Haonan Bian.
\newblock {LLM}-empowered knowledge graph construction: A survey.
\newblock \emph{arXiv preprint arXiv:2510.20345}, 2025.

\bibitem[{Braintrust Data, Inc.}(2026)]{braintrust}
{Braintrust Data, Inc.}
\newblock {Braintrust}: Evaluation and observability platform for {AI} applications.
\newblock \url{https://www.braintrust.dev}, 2026.

\bibitem[Dong et~al.(2014)Dong, Gabrilovich, Heitz, Horn, Lao, Murphy, Strohmann, Sun, and Zhang]{dong2014knowledgevault}
Xin Dong, Evgeniy Gabrilovich, Geremy Heitz, Wilko Horn, Ni~Lao, Kevin Murphy, Thomas Strohmann, Shaohua Sun, and Wei Zhang.
\newblock Knowledge vault: A web-scale approach to probabilistic knowledge fusion.
\newblock In \emph{Proceedings of the 20th {ACM} {SIGKDD} International Conference on Knowledge Discovery and Data Mining (KDD)}, pages 601--610, 2014.

\bibitem[Elkan and Noto(2008)]{elkan2008pu}
Charles Elkan and Keith Noto.
\newblock Learning classifiers from only positive and unlabeled data.
\newblock In \emph{Proceedings of the 14th {ACM} {SIGKDD} International Conference on Knowledge Discovery and Data Mining (KDD)}, pages 213--220, 2008.

\bibitem[{Firecrawl}(2026)]{firecrawl}
{Firecrawl}.
\newblock {Firecrawl}: Web crawling and scraping {API} for {LLM} applications.
\newblock \url{https://firecrawl.dev}, 2026.

\bibitem[Gao et~al.(2023{\natexlab{a}})Gao, Dai, Pasupat, Chen, Chaganty, Fan, Zhao, Lao, Lee, Juan, and Guu]{gao2023rarr}
Luyu Gao, Zhuyun Dai, Panupong Pasupat, Anthony Chen, Arun~Tejasvi Chaganty, Yicheng Fan, Vincent Zhao, Ni~Lao, Hongrae Lee, Da-Cheng Juan, and Kelvin Guu.
\newblock {RARR}: Researching and revising what language models say, using language models.
\newblock In \emph{Proceedings of the 61st Annual Meeting of the Association for Computational Linguistics (ACL)}, pages 16477--16508, 2023{\natexlab{a}}.

\bibitem[Gao et~al.(2023{\natexlab{b}})Gao, Yen, Yu, and Chen]{gao2023alce}
Tianyu Gao, Howard Yen, Jiatong Yu, and Danqi Chen.
\newblock Enabling large language models to generate text with citations.
\newblock In \emph{Proceedings of the 2023 Conference on Empirical Methods in Natural Language Processing (EMNLP)}, 2023{\natexlab{b}}.

\bibitem[Guo et~al.(2017)Guo, Pleiss, Sun, and Weinberger]{guo2017calibration}
Chuan Guo, Geoff Pleiss, Yu~Sun, and Kilian~Q. Weinberger.
\newblock On calibration of modern neural networks.
\newblock In \emph{Proceedings of the 34th International Conference on Machine Learning (ICML)}, pages 1321--1330, 2017.

\bibitem[Huang et~al.(2025)Huang, Chen, Sheng, Li, and Zhang]{huang2025graphjudge}
Haoyu Huang, Chong Chen, Zeang Sheng, Yang Li, and Wentao Zhang.
\newblock Can {LLMs} be good graph judge for knowledge graph construction?
\newblock In \emph{Proceedings of the 2025 Conference on Empirical Methods in Natural Language Processing (EMNLP)}, pages 10940--10959, 2025.

\bibitem[Ji et~al.(2023)Ji, Lee, Frieske, Yu, Su, Xu, Ishii, Bang, Madotto, and Fung]{ji2023hallucination}
Ziwei Ji, Nayeon Lee, Rita Frieske, Tiezheng Yu, Dan Su, Yan Xu, Etsuko Ishii, Ye~Jin Bang, Andrea Madotto, and Pascale Fung.
\newblock Survey of hallucination in natural language generation.
\newblock \emph{ACM Computing Surveys}, 55\penalty0 (12), 2023.

\bibitem[{LangChain}(2024)]{langgraph}
{LangChain}.
\newblock {LangGraph}: A low-level orchestration framework for stateful agents.
\newblock \url{https://github.com/langchain-ai/langgraph}, 2024.

\bibitem[Lewis et~al.(2020)Lewis, Perez, Piktus, Petroni, Karpukhin, Goyal, K{\"u}ttler, Lewis, Yih, Rockt{\"a}schel, Riedel, and Kiela]{lewis2020rag}
Patrick Lewis, Ethan Perez, Aleksandra Piktus, Fabio Petroni, Vladimir Karpukhin, Naman Goyal, Heinrich K{\"u}ttler, Mike Lewis, Wen-tau Yih, Tim Rockt{\"a}schel, Sebastian Riedel, and Douwe Kiela.
\newblock Retrieval-augmented generation for knowledge-intensive {NLP} tasks.
\newblock In \emph{Advances in Neural Information Processing Systems (NeurIPS)}, 2020.

\bibitem[Liu et~al.(2024)Liu, Lin, Hewitt, Paranjape, Bevilacqua, Petroni, and Liang]{liu2024lost}
Nelson~F. Liu, Kevin Lin, John Hewitt, Ashwin Paranjape, Michele Bevilacqua, Fabio Petroni, and Percy Liang.
\newblock Lost in the middle: How language models use long contexts.
\newblock \emph{Transactions of the Association for Computational Linguistics}, 12:\penalty0 157--173, 2024.

\bibitem[Min et~al.(2023)Min, Krishna, Lyu, Lewis, Yih, Koh, Iyyer, Zettlemoyer, and Hajishirzi]{min2023factscore}
Sewon Min, Kalpesh Krishna, Xinxi Lyu, Mike Lewis, Wen-tau Yih, Pang~Wei Koh, Mohit Iyyer, Luke Zettlemoyer, and Hannaneh Hajishirzi.
\newblock {FActScore}: Fine-grained atomic evaluation of factual precision in long form text generation.
\newblock In \emph{Proceedings of the 2023 Conference on Empirical Methods in Natural Language Processing (EMNLP)}, 2023.

\bibitem[Mintz et~al.(2009)Mintz, Bills, Snow, and Jurafsky]{mintz2009distant}
Mike Mintz, Steven Bills, Rion Snow, and Dan Jurafsky.
\newblock Distant supervision for relation extraction without labeled data.
\newblock In \emph{Proceedings of the Joint Conference of the 47th Annual Meeting of the {ACL} and the 4th International Joint Conference on Natural Language Processing of the {AFNLP}}, pages 1003--1011, 2009.

\bibitem[Nakano et~al.(2021)Nakano, Hilton, Balaji, Wu, Ouyang, Kim, Hesse, Jain, Kosaraju, Saunders, Jiang, Cobbe, Eloundou, Krueger, Button, Knight, Chess, and Schulman]{nakano2021webgpt}
Reiichiro Nakano, Jacob Hilton, Suchir Balaji, Jeff Wu, Long Ouyang, Christina Kim, Christopher Hesse, Shantanu Jain, Vineet Kosaraju, William Saunders, Xu~Jiang, Karl Cobbe, Tyna Eloundou, Gretchen Krueger, Kevin Button, Matthew Knight, Benjamin Chess, and John Schulman.
\newblock {WebGPT}: Browser-assisted question-answering with human feedback.
\newblock \emph{arXiv preprint arXiv:2112.09332}, 2021.

\bibitem[{Neo4j, Inc.}(2026)]{neo4j}
{Neo4j, Inc.}
\newblock {Neo4j} graph database.
\newblock \url{https://neo4j.com}, 2026.

\bibitem[Northcutt et~al.(2021)Northcutt, Athalye, and Mueller]{northcutt2021pervasive}
Curtis~G. Northcutt, Anish Athalye, and Jonas Mueller.
\newblock Pervasive label errors in test sets destabilize machine learning benchmarks.
\newblock In \emph{Advances in Neural Information Processing Systems (NeurIPS), Datasets and Benchmarks Track}, 2021.

\bibitem[{OpenRouter}(2026)]{openrouter}
{OpenRouter}.
\newblock {OpenRouter}: Unified {API} gateway for language models.
\newblock \url{https://openrouter.ai}, 2026.

\bibitem[Pan et~al.(2024)Pan, Luo, Wang, Chen, Wang, and Wu]{pan2024unifying}
Shirui Pan, Linhao Luo, Yufei Wang, Chen Chen, Jiapu Wang, and Xindong Wu.
\newblock Unifying large language models and knowledge graphs: A roadmap.
\newblock \emph{IEEE Transactions on Knowledge and Data Engineering}, 36\penalty0 (7):\penalty0 3580--3599, 2024.

\bibitem[Raji et~al.(2020)Raji, Smart, White, Mitchell, Gebru, Hutchinson, Smith-Loud, Theron, and Barnes]{raji2020closing}
Inioluwa~Deborah Raji, Andrew Smart, Rebecca~N. White, Margaret Mitchell, Timnit Gebru, Ben Hutchinson, Jamila Smith-Loud, Daniel Theron, and Parker Barnes.
\newblock Closing the {AI} accountability gap: Defining an end-to-end framework for internal algorithmic auditing.
\newblock In \emph{Proceedings of the 2020 Conference on Fairness, Accountability, and Transparency (FAT* '20)}, pages 33--44, 2020.

\bibitem[Rashkin et~al.(2023)Rashkin, Nikolaev, Lamm, Aroyo, Collins, Das, Petrov, Tomar, Turc, and Reitter]{rashkin2023attribution}
Hannah Rashkin, Vitaly Nikolaev, Matthew Lamm, Lora Aroyo, Michael Collins, Dipanjan Das, Slav Petrov, Gaurav~Singh Tomar, Iulia Turc, and David Reitter.
\newblock Measuring attribution in natural language generation models.
\newblock \emph{Computational Linguistics}, 49\penalty0 (4):\penalty0 777--840, 2023.

\bibitem[Rasmussen et~al.(2025)Rasmussen, Paliychuk, Beauvais, Ryan, and Chalef]{rasmussen2025zep}
Preston Rasmussen, Pavlo Paliychuk, Travis Beauvais, Jack Ryan, and Daniel Chalef.
\newblock {Zep}: A temporal knowledge graph architecture for agent memory.
\newblock \emph{arXiv preprint arXiv:2501.13956}, 2025.

\bibitem[Sculley et~al.(2015)Sculley, Holt, Golovin, Davydov, Phillips, Ebner, Chaudhary, Young, Crespo, and Dennison]{sculley2015debt}
D.~Sculley, Gary Holt, Daniel Golovin, Eugene Davydov, Todd Phillips, Dietmar Ebner, Vinay Chaudhary, Michael Young, Jean-Fran{\c{c}}ois Crespo, and Dan Dennison.
\newblock Hidden technical debt in machine learning systems.
\newblock In \emph{Advances in Neural Information Processing Systems (NIPS)}, pages 2503--2511, 2015.

\bibitem[Wadhwa et~al.(2023)Wadhwa, Amir, and Wallace]{wadhwa2023revisiting}
Somin Wadhwa, Silvio Amir, and Byron~C. Wallace.
\newblock Revisiting relation extraction in the era of large language models.
\newblock In \emph{Proceedings of the 61st Annual Meeting of the Association for Computational Linguistics (ACL)}, 2023.

\bibitem[Wang et~al.(2023)Wang, Wei, Schuurmans, Le, Chi, Narang, Chowdhery, and Zhou]{wang2023selfconsistency}
Xuezhi Wang, Jason Wei, Dale Schuurmans, Quoc Le, Ed~Chi, Sharan Narang, Aakanksha Chowdhery, and Denny Zhou.
\newblock Self-consistency improves chain of thought reasoning in language models.
\newblock In \emph{International Conference on Learning Representations (ICLR)}, 2023.

\bibitem[Yao et~al.(2023)Yao, Zhao, Yu, Du, Shafran, Narasimhan, and Cao]{yao2023react}
Shunyu Yao, Jeffrey Zhao, Dian Yu, Nan Du, Izhak Shafran, Karthik Narasimhan, and Yuan Cao.
\newblock {ReAct}: Synergizing reasoning and acting in language models.
\newblock In \emph{International Conference on Learning Representations (ICLR)}, 2023.

\bibitem[Zheng et~al.(2023)Zheng, Chiang, Sheng, Zhuang, Wu, Zhuang, Lin, Li, Li, Xing, Zhang, Gonzalez, and Stoica]{zheng2023judging}
Lianmin Zheng, Wei-Lin Chiang, Ying Sheng, Siyuan Zhuang, Zhanghao Wu, Yonghao Zhuang, Zi~Lin, Zhuohan Li, Dacheng Li, Eric~P. Xing, Hao Zhang, Joseph~E. Gonzalez, and Ion Stoica.
\newblock Judging {LLM}-as-a-judge with {MT-Bench} and {Chatbot Arena}.
\newblock In \emph{Advances in Neural Information Processing Systems (NeurIPS), Datasets and Benchmarks Track}, 2023.

\end{thebibliography}

% ===========================================================================
\clearpage
\appendix

\section{Implementation Details}
\label{app:impl}

\paragraph{Orchestration.} The pipeline consists of two LangGraph state machines. The outer graph maps and fetches the vendor's site, plans and runs extraction over the retained pages, then starts one discovery agent per relation type and sends the pages they keep back through planning and extraction once. The inner graph handles one page: it loads the cached page, extracts candidates, resolves their endpoints, runs grounding, writes the surviving edges and records the outcome. All model calls go through OpenRouter.

\paragraph{Prompt versions.} Every prompt is loaded at run time from Braintrust by name, so wording can change without a code deployment and every version is kept. An extractor version hash covers the model name, the output schema, the grounding policy and the name and version of every extraction, referent and panel prompt. Extraction is keyed on (target, page content, relation type, extractor version), so changing any of these prompts re-runs extraction, while an unchanged pipeline never processes the same page twice.

\paragraph{Vendor-site crawl.} URLs are normalised (lower-case host, no tracking parameters, no fragment) and excluded when they point to static assets, login pages, calendar crawler traps, \code{developer.*} documentation hosts, sitemaps or \code{robots.txt}. Pages are scraped as main-content Markdown, with one retry after rendering when the content is a loading placeholder or shorter than 200 characters. Every page is stored once, under an identifier derived from its URL and content.

\paragraph{Discovery agent.} The agent is built with the \code{deepagents} library. Tool budgets are reserved atomically so that parallel tool calls cannot exceed them. Before the agent starts, one fixed opening search is run. The pages the agent keeps are checked against the URLs it actually observed, and the selection is topped up deterministically with observed pages whose title or search snippet names the product.

\paragraph{Entity resolution.} Names that do not match a known entity on the page exactly are embedded and compared with existing entities of the same type, and Graphiti's resolver then tries normalised names, MinHash fuzzy matching and a model call.

\paragraph{Persistence.} Each edge also stores the page's content hash and a text embedding of the quote, and each page is kept as an episode node with its full content. Edges are written inactive and activated together once the extraction completes; a later revision deactivates an earlier edge only when it reproduces the same relation.

\clearpage
\section{Grounding Checks and Criteria}
\label{app:checks}

Table~\ref{tab:checks} lists the deterministic shape checks, and Table~\ref{tab:criteria} the nine criteria of the semantic panel.

\begin{table}[H]
\centering
\small
\begin{tabularx}{\linewidth}{lX}
\toprule
Check & Rejects \\
\midrule
\code{incomplete\_relationship} & a missing object, object type or relation \\
\code{unnamed\_subject} / \code{unnamed\_object} & a generic endpoint (``hospitals'', ``EHR systems'', ``the platform'') \\
\code{subject\_is\_individual} & a person (title or clinical credential) given as the organisation \\
\code{meta\_commentary\_quote} & a quote that is the model's own arrow notation, not page text \\
\code{quote\_not\_in\_source} & a quote less than 75\% of whose characters match the page \\
\code{quote\_not\_contiguous} & a quote whose longest contiguous match is under 95\% (pre-empted by splice repair) \\
\code{target\_not\_endpoint} & a relation that does not involve the target product \\
\code{self\_edge} & identical subject and object \\
\code{profile\_shape\_mismatch} & an edge shape the relation type does not allow \\
\code{target\_as\_customer\_subject} & the target product named as its own customer \\
\code{target\_as\_integration\_partner} & the vendor's own name fused to the partner (``Care-Wellframe'') \\
\code{subject/object\_named\_only\_in\_url} & a name that appears only inside a link's URL, not in visible text \\
\bottomrule
\end{tabularx}
\caption{Deterministic shape checks. Two further codes (an endpoint absent from the evidence) exist but are disabled for both relation types.}
\label{tab:checks}
\end{table}

\begin{table}[H]
\centering
\small
\begin{tabularx}{\linewidth}{lXc}
\toprule
Criterion & True when & Gates \\
\midrule
\code{is\_named\_organization} & the subject is a specific, real, named organisation, not a person, a category, a placeholder (``your practice''), an anonymised description, or the vendor's own brand & Adopts \\
\code{relationship\_actually\_happened} & the relation is stated as a completed fact, not a plan, a hope, or something cancelled & both \\
\code{relationship\_claim\_supported} & the quote states this type of relation between these entities, not a co-mention, an investment, an award or a network membership & both \\
\code{is\_third\_party\_vendor} & the partner is independent of the vendor, not its own product, sub-brand or acquisition & Integrates \\
\code{is\_customer\_facing\_integration} & the integration is one a customer would use, not swappable infrastructure (a cloud host, a generic model provider, a coding assistant) & Integrates \\
\code{is\_specific\_named\_entity} & the partner is a named product, not a category (``EHRs'', ``practice management system'') & Integrates \\
\code{page\_frames\_them\_as\_users} & the page itself presents the listed organisations as users, rather than as a consultancy's clients, sponsors or award winners & Adopts \\
\code{subject\_is\_the\_claimant} & the claim is not a third party's first-person statement in which the subject is merely listed & Integrates \\
\code{relationship\_shape\_matches\_profile} & the relation type matches how the quote describes it (a practice adopting a product is not an integration) & both \\
\bottomrule
\end{tabularx}
\caption{The semantic panel's nine criteria, abridged from their definitions in the output schema. Every call answers all nine; each passes when at least half of the calls that returned say yes.}
\label{tab:criteria}
\end{table}

\section{Prompts}
\label{app:prompts}

Pipeline prompts are stored in Braintrust and versioned there. We reproduce the two judge prompts used to build the \solkey{} and the referent prompt for customer relations on vendor pages; the remaining prompts are identified by slug in Table~\ref{tab:slugs}. Placeholders are in braces.

\subsection{Judge Prompt for Customer Relations}
\begin{Verbatim}[fontsize=\scriptsize, frame=single]
An organisation: {subject}
A product: {object}

Does the page below say that this organisation uses this product?

Adoption means the organisation actually uses it or is being provided it --
deployed, implemented, rolled out, live on it, selected and now running it, or
co-developed it with the vendor and uses the result.

It does not count when:
- the relationship is an investment, an award, sponsorship, membership of a
  network or programme, or a research study with no use implied
- the organisation named is a different one of a similar name
- the only named party is an individual, with no hospital, practice or health
  system tied to them
- the page is a careers page, an investor page or press boilerplate describing a
  company rather than who uses its products
- use is planned or announced rather than in place

Answer with one word, YES or NO, then one sentence of justification.

PAGE:
{page}
\end{Verbatim}

\subsection{Judge Prompt for Integration Relations}
\begin{Verbatim}[fontsize=\scriptsize, frame=single]
Two products: {subject} and {object}.

Does the page below state that these two are connected?

It counts as a connection only if the page describes one of them being integrated
with, connected to, embedded in, interoperable with, or combined with the other --
a real technical connection a customer would use.

It does not count when:
- the claim is generic compatibility that never names this specific product
  ("works with most EHRs", "integrates with any PMS")
- it is an investment, acquisition, reseller or commercial partnership with no
  technical connection stated
- one of them is general-purpose infrastructure the other merely builds on -- a
  cloud host, an LLM provider, a database -- which could be swapped for an
  equivalent without changing what the product does
- either side is a category rather than a named product ("practice management
  system", "EHRs")
- the connection is planned, intended or announced rather than in place
- the two merely appear together in a list, a logo wall, or the same sentence

Answer with one word, YES or NO, then one sentence of justification.

PAGE:
{page}
\end{Verbatim}

\subsection{Referent Prompt for Customer Relations on Vendor Pages (Abridged)}
The full prompt adds two worked examples and paragraphs on press releases, directories and investor pages. Its output field \code{integrated} holds the yes/no verdict (the name is shared with the integration prompts).
\begin{Verbatim}[fontsize=\scriptsize, frame=single]
Look at this organisation: {{subject}}
Look at this product: {{object}}

Adoption means the organisation actually uses the product, or is being provided
it -- deployed, implemented, rolled out, live on it, selected and now running it,
or co-developed it with the vendor and uses the result. It is not an investment,
an award, sponsorship, membership of a network or programme, or a research study
with no use implied.

Be careful whose page this is. The page is {{source_url}}. Many pages belong to a
third company listing the organisations it works with, or the products it
resells. An organisation and a product both appearing on such a page does not
mean that organisation uses that product. First-person words like "we" and "our"
belong to the page's owner. Only answer yes if the page says this organisation
uses this product.

[... exact organisation, exact product, individuals, page types ...]

On the vendor's own site the page's structure can establish adoption as plainly
as a sentence does. A case study, a customer spotlight, a "see how X did Y"
teaser, or a heading naming an organisation above an account of what it
achieved -- each of these is the vendor presenting that organisation as a
customer, and the body does not have to repeat the product's name for it to
count. [...] What does not count is a bare logo or a name sitting in a grid with
nothing said about it.

The test: would someone at this organisation read this page and recognise it as
describing their own use of the product?

[... two worked examples ...]

Give your answer in three fields, in this order: `reason`, your argument in a
sentence or two; `integrated`, yes or no; `proof_sentence`, the sentence from the
page that shows it, copied word for word, or empty if there is none.

PAGE:
{{page_content}}
\end{Verbatim}

\begin{table}[H]
\centering
\small
\begin{tabularx}{\linewidth}{@{}lX@{}}
\toprule
Step & Braintrust slug(s) \\
\midrule
Extraction & \code{customer-extraction-8f90}\newline\code{integration-extraction-4de1} \\
Referent, third-party pages & \code{referent-solution-customers-7f21}\newline\code{referent-solution-0cd9} \\
Referent, vendor pages & \code{referent-solution-customers-firstparty-1eb4}\newline\code{referent-solution-integrations-firstparty-71e0} \\
Semantic panel & \code{semantic-grounding-check-customers-fe3b}\newline\code{semantic-grounding-check-integrations-a12e} \\
Discovery agent & \code{discovery-system-prompt-32a6}\newline\code{customer-discovery-f006}\newline\code{integration-discovery-c2a2} \\
\bottomrule
\end{tabularx}
\caption{Prompt slugs. The extractor version hash covers the extraction, referent and semantic-panel prompts and their versions.}
\label{tab:slugs}
\end{table}

\section{Confidence Score Details}
\label{app:score}

\paragraph{Signals.} Each relation has 22 signals. Eight are vote shares: for eight of the nine panel criteria, the fraction of the five calls that said yes (the shape criterion was not included). Eleven are flags: whether the panel ran, whether the referent was asked, whether it confirmed, whether the quote names both sides, whether the quote is contiguous, whether the subject or the object appears in the URL, whether the page is the vendor's own, whether the relation is a customer relation, whether the edge type is allowed, and whether a shape check rejected it. Three are numbers: the quote length, how often the quote is reused on the page, and how many relations the extractor emitted in the same call.

\paragraph{Fitting and search.} The score is a logistic regression with light ridge regularisation. All results are out-of-sample under five-fold cross-validation; in the subset search, feature standardisation is computed on each training fold alone. We evaluated every subset of up to eight features (603,500 subsets), extended the forty best eight-feature subsets by beam search up to all 22 features, and separately searched 141,724 decision trees with up to five features and depth up to four. The search took about two hours of compute and no model calls.

\paragraph{Results.} The best F1 is 0.907, reached with thirteen features; the best five-feature model reaches 0.898, and the best tree 0.900. These values are optimistic, because each model was chosen from many candidates using the same folds that score it. Figure~\ref{fig:pr} uses all 22 features, with no selection. The tree reads as a rule: relations with a unanimous panel are 0.89 to 0.97 correct, and when the panel is split, those the referent confirmed are 0.89 correct and those it refused are 0.33. Being asked by the referent carries a negative weight where it appears, since the referent runs only when the quote cannot settle the relation alone. The customer-relation flag is the heaviest weight in most fitted models.

\section{Reproducibility}
\label{app:repro}

The public repository (see Code Availability) holds the pipeline, adapted prompts, a ten-product sample and public editions of the experiment write-ups, but not the full-corpus data. The pipeline, scoring scripts and experiment write-ups used for this paper are held in the industry partner's private repository; the corpus table is not in that repository because of its size. The quantities in this paper come from the following artefacts:
\begin{itemize}
  \item the corpus table of 6,429 candidates with every vote, referent decision and rejection reason, its run metadata (model and prompt versions), and the outcomes of the re-run candidates;
  \item the judge's verdicts and the review (6,428 entries, with a written reason for each of the 911 corrections);
  \item the scorer for both benchmarks and the three confidence-score scripts (fitting, feature-subset search and tree search);
  \item a snapshot of the golden reference (61 products, 749 customer and 249 integration rows).
\end{itemize}

\end{document}